\documentclass[journal]{IEEEtran}
\usepackage{balance}

\usepackage{amsmath,amssymb,amsfonts}
\usepackage{algorithmic}
\usepackage[hidelinks]{hyperref}
\usepackage[linesnumbered,ruled,vlined]{algorithm2e}
\usepackage{graphicx}
\usepackage{textcomp}
\usepackage{xcolor}
\usepackage{caption}
\usepackage{subcaption}
\usepackage{multirow}
\usepackage{enumitem}
\ifCLASSINFOpdf
\else
\fi
\usepackage{array}
\newcolumntype{M}[1]{>{\centering\arraybackslash}m{#1}}

\begin{document}

\title{Deployment of UAVs for Optimal Multihop Ad-hoc Networks Using Particle Swarm Optimization and Behavior-based Control}

\author{Ngan Duong Thi Thuy, Duy Nam Bui, Manh Duong Phung and Hung Pham Duy
\thanks{Ngan Duong Thi Thuy, Duy Nam Bui and Hung Pham Duy are with the VNU University of Engineering and Technology, Hanoi, Vietnam. {\fontfamily{qcr}\selectfont
ngan.uet@gmail.com; duynam.robotics@gmail.com; hungpd@vnu.edu.vn}}

\thanks{Manh Duong Phung is with the Fulbright University Vietnam, Ho Chi Minh City, Vietnam. {\fontfamily{qcr}\selectfont
duong.phung@fulbright.edu.vn}}

\thanks{This research has been done under the research project QG.21.26 ``Resilience for Preserving Multi-Robot Network in Dynamic Environment'' of Vietnam National University, Hanoi.}
\vspace{0.3cm}
}


%



\maketitle

\begin{abstract}
This study proposes an approach for establishing an optimal multihop ad-hoc network using multiple unmanned aerial vehicles (UAVs) to provide emergency communication in disaster areas. The approach includes two stages, one uses particle swarm optimization (PSO) to find optimal positions to deploy UAVs, and the other uses a behavior-based controller to navigate the UAVs to their assigned positions without colliding with obstacles in an unknown environment. Several constraints related to the UAVs' sensing and communication ranges have been imposed to ensure the applicability of the proposed approach in real-world scenarios. A number of simulation experiments with data loaded from real environments have been conducted. The results show that our proposed approach is not only successful in establishing multihop ad-hoc routes but also meets the requirements for real-time deployment of UAVs. 
\end{abstract}


\IEEEpeerreviewmaketitle
\section{Introduction}

When natural disasters happen, it is critical to maintain communication channels between the affected areas and rescue coordination centers to effectively coordinate search and rescue operations. An emergency wireless network is typically needed as existing communication infrastructure is either damaged or destroyed during the disaster. Ad-hoc networks provide a relevant solution in this scenario since they are reliable in harsh conditions and can be deployed quickly without requiring existing communication infrastructure \cite{doi:10.1155/2015/647037,6710069}. The use of these networks has been discussed in \cite{anjum2017review,5730373,8167124} where satellites, mobile devices, and remaining active base stations are utilized to form the networks. However, the flexibility and scalability of these networks are still limited due to their dependence on remaining active devices, which are hard to manage during a disaster.  

Recently, unmanned aerial vehicles (UAVs) have shown their potential to search and rescue applications thanks to their mobility in working environments and capability of carrying different types of sensors \cite{8756125, 9214446, PHUNG2020106705}. Several works have been conducted to address issues related to UAV-based network communication such as routing protocols \cite{maxa2015secure}, data authentication schemes \cite{sun2020data}, topology control \cite{zhao2012topology}, and medium access control \cite{chua2012medium}. However, the location allocation and routing problems for UAVs in unknown environments are hardly addressed. When establishing a multihop ad-hoc route, it is necessary to optimize the position of UAV nodes for efficient signal transmission. The route to navigate each UAV to its goal position should also be determined to ensure safe operation of the UAV. 

In this work, we propose a new approach to deploying multiple UAVs to establish an ad-hoc network in unknown environments. First, we search for the optimal position of each UAV by formulating it as an optimization problem and then use particle swarm optimization (PSO) to solve it. A behavior-based target tracking control is then proposed to navigate the UAVs to their optimal positions while avoiding obstacles. Simulation results show that the proposed approach can establish a communication channel between the base station and the disaster area in complex environments. The computed positions and routes are also safe and efficient for UAV operation.
\section{Problem formulation}
\label{sec:pro}
\begin{figure}
    \centering
    \begin{subfigure}[b]{0.48\textwidth}
    \centering
    \includegraphics[width=\textwidth]{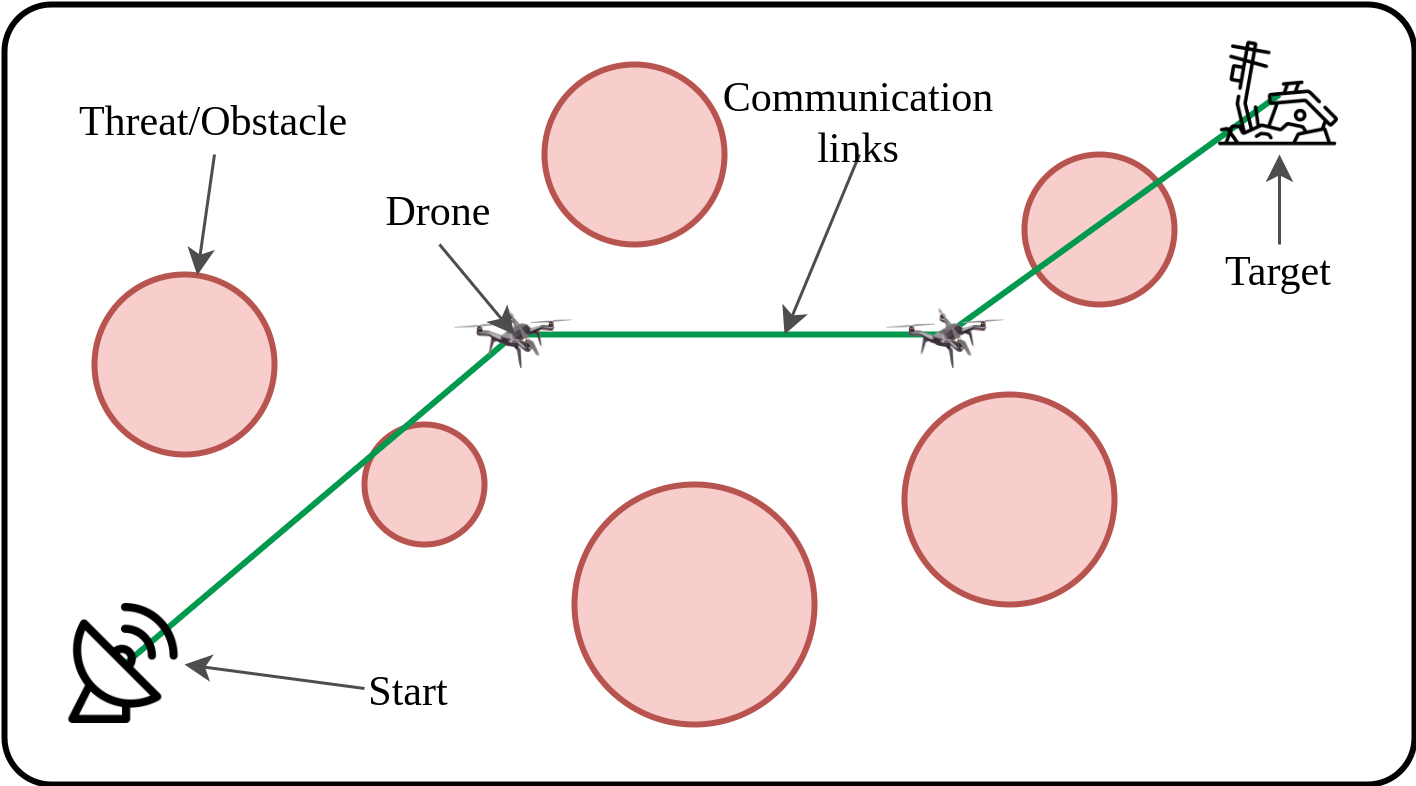}
    \caption{Creation of a multihop ad-hoc route}
    \label{fig:modela}
    \end{subfigure}
    \begin{subfigure}[b]{0.5\textwidth}
    \centering
    \includegraphics[width=0.55\textwidth]{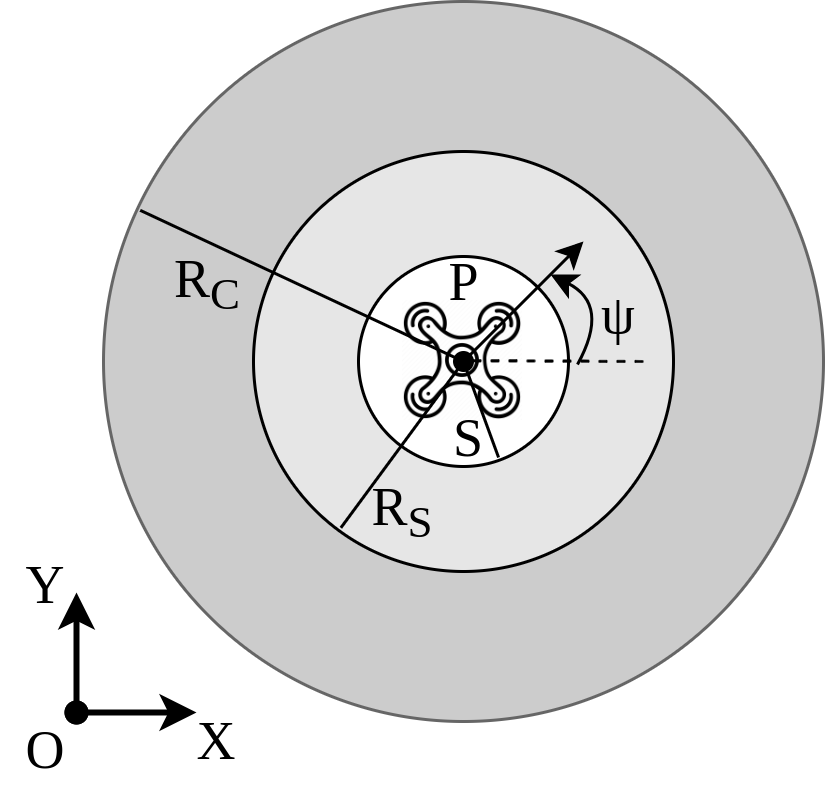}
    \caption{The UAV model with its sensing and communication ranges}
    \label{fig:modelb}
    \end{subfigure}
    \caption{The multihop ad-hoc network and UAV model.}
    \label{fig:pro}
\end{figure}
\begin{figure}
    \centering
    \includegraphics[width=0.48\textwidth]{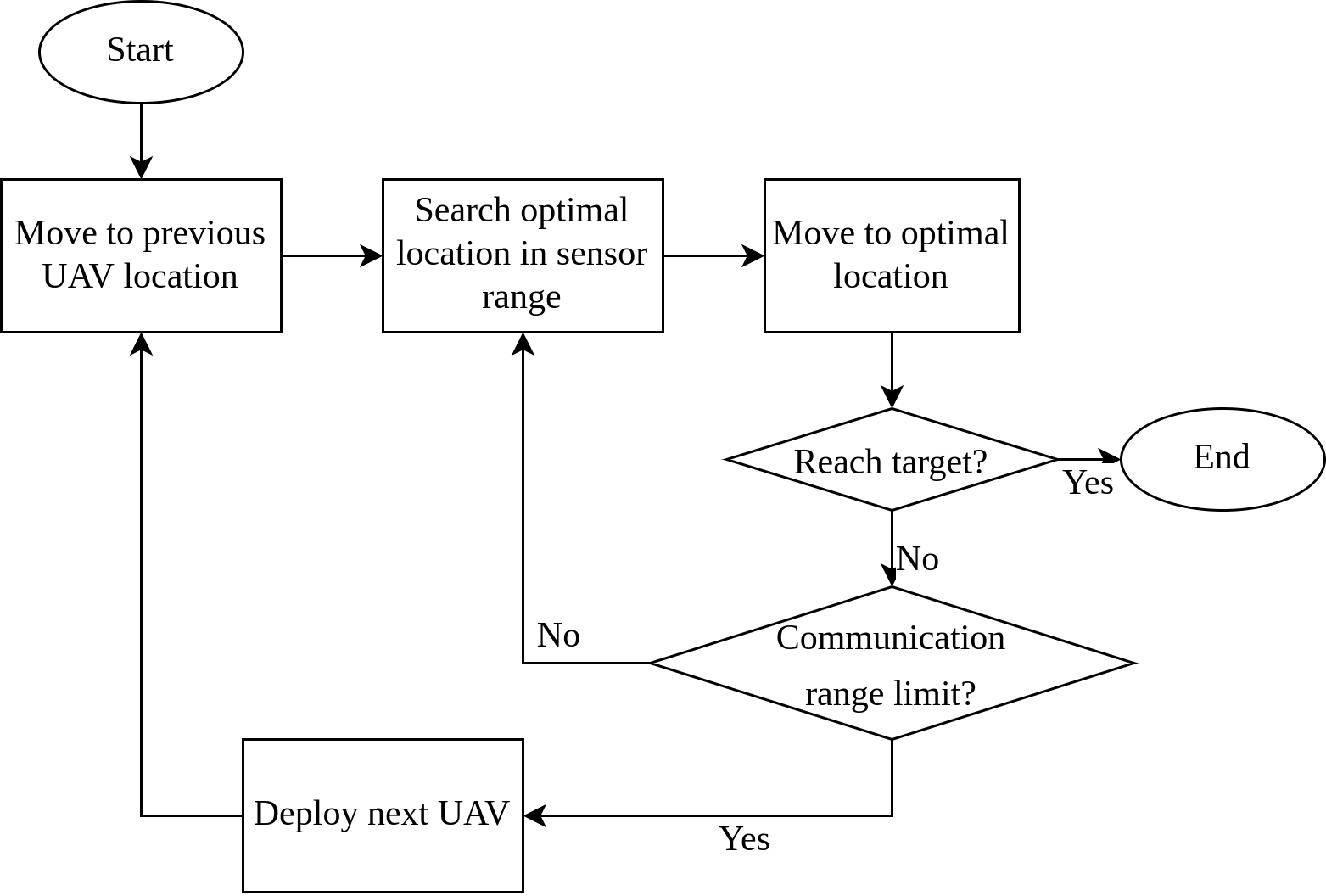}
    \caption{Deployment approach diagram}
    \label{fig:approach}
\end{figure}
This work focuses on establishing an optimal multihop ad-hoc route that connects a base station to a disaster area via a sequence of UAV nodes, as illustrated in Fig. \ref{fig:modela}. The UAVs act as intermediate routers to relay messages. Each UAV is equipped with a GPS module and an inertial measurement unit (IMU) to determine its position and orientation with respect to a global coordinate system. We assume that the UAV has a sensory system that can detect obstacles and a network module that is capable of carrying out peer-to-peer communication with other UAVs within its range. We also assume that the number of UAVs is sufficient and the disaster area's location is known. 

\subsection{UAV model and constraints}

The UAV is modelled as a particle with position $P=\left[x,y,z\right]^T$ and heading angle $\psi$, as illustrated in Fig. \ref{fig:modelb}. The safe zone of the UAV is represented as the area surrounded by a circle with a radius of $S$. The communication and sensing ranges are represented by circles with the radii of $R_C$ and $R_S$, respectively. Noting that the sensing range is typically much small than the communication. Let $V=\left[v_x,v_y,v_z\right]^T$ be the UAV's velocity. The position and orientation of the UAV after each time step $T$ are updated as follows:

\begin{equation}
    \begin{aligned}
    P\left(t+T\right)&=P\left(t\right)+V\left(t\right)T\\
    \psi\left(t+T\right)&=\text{atan2}\left(v_{y},v_{x}\right)
    \end{aligned}
\end{equation}

To form an ad-hoc network, the following constraints must be satisfied:
\begin{enumerate}[label=(\roman*)]
     \item Each UAV stays within the communication range of its adjacent UAVs.
     \item Each UAV maintains a safe distance from obstacles and a suitable altitude above the ground during its operation.
\end{enumerate}

\subsection{Deployment approach}

In this work, our approach to deploying UAVs is illustrated in Fig. \ref{fig:approach}. UAVs are being deployed sequentially in which each UAV concurrently searches for the next optimal position within its sensing range until reaching the communication boundary. The found position closest to the communication boundary is then chosen as the final position for that UAV. After that, the next UAV will be dispatched to first go to the last found optimal position and then carry out the search process to find its optimal location. The process is repeated until the target is reached. We use PSO to search for optimal positions and a behavior-based controller to navigate the UAVs to those positions.

\section{PSO-based optimal position identification}
\label{sec:sea}
This section describes our approach to find optimal positions for UAVs to establish a multihop ad-hoc communication route from the base station to the disaster area. 

\subsection{Fitness function definition}
The search process is formulated as an optimization problem that recursively looks for the optimal intermediate point $P_{inew}$ within the UAV's sensing range $R_S$ until reaching its communication limit $R_C$. The search for $P_{inew}$ is carried out by fulfilling the following criteria: 1) minimizing the angle between the UAV and the target, 2) maintaining the UAV's location within the proximity of its neighbors' communication range, 3) avoiding obstacles. Hence, the fitness function is designed in the form:

\begin{equation}
    F=\sum_{k=1}^{4}b_{k}F_{k}
    \label{eqn:cost}
\end{equation}
where $b_k$ is the weight coefficient and $F_k$ is the cost associated to search criterion $k$. In this work, we define $F_1$ for obstacle avoidance, $F_2$ for target angle minimization and $F_3$ and $F_4$ for optimizing the sensing and communication links.

\subsubsection{Avoiding obstacles}
\begin{figure}
    \centering
    \includegraphics[width=0.35\textwidth]{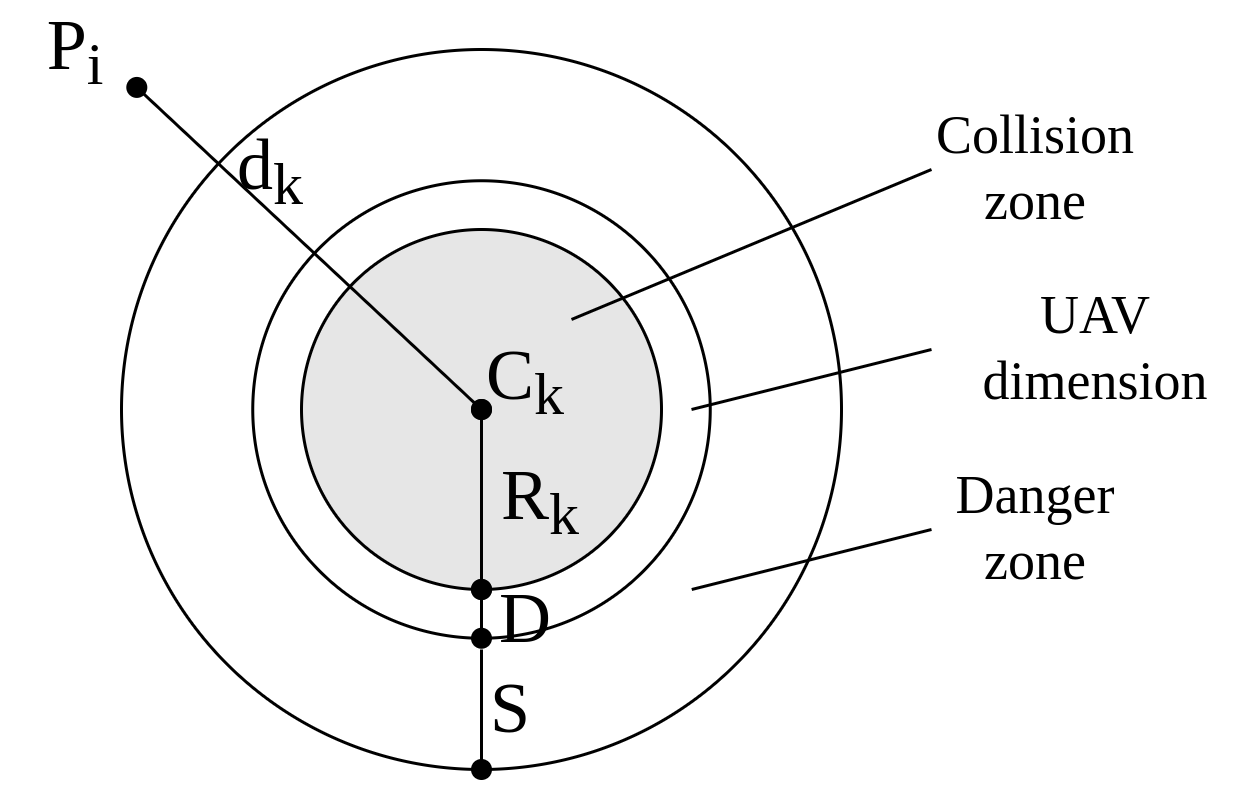}
    \caption{Obstacle avoidance illustration}
    \label{fig:avoid_threats}
\end{figure}

Assume that the UAV has a size of $D$. At height $z$, an obstacle can be represented by a circle with center $C_k$ and radius $R_k$. To avoid obstacles, the UAV needs to maintain a safe distance $S$ to the obstacles' boundary as illustrated in Fig. \ref{fig:avoid_threats}. Let $P_i$ be the location of UAV $i$ and $d_k$ be the distance between $P_i$ and $C_k$. The objective function for obstacle avoidance is defined as follows:
\begin{equation}
    F_{1}=\left\{ \begin{array}{cc}
    0 & \text{if } d_{k}>R_{k}+D+S\\
    R_{k}+D+S-d_{k} &\text{if } R_{k}+D<d_{k}\leq R_{k}+D+S\\
    \infty &\text{if } d_{k}\leq R_{k}+D
    \end{array}\right.
\end{equation}

\subsubsection{Minimizing the target angle}
\begin{figure}
    \centering
    \includegraphics[width=0.3\textwidth]{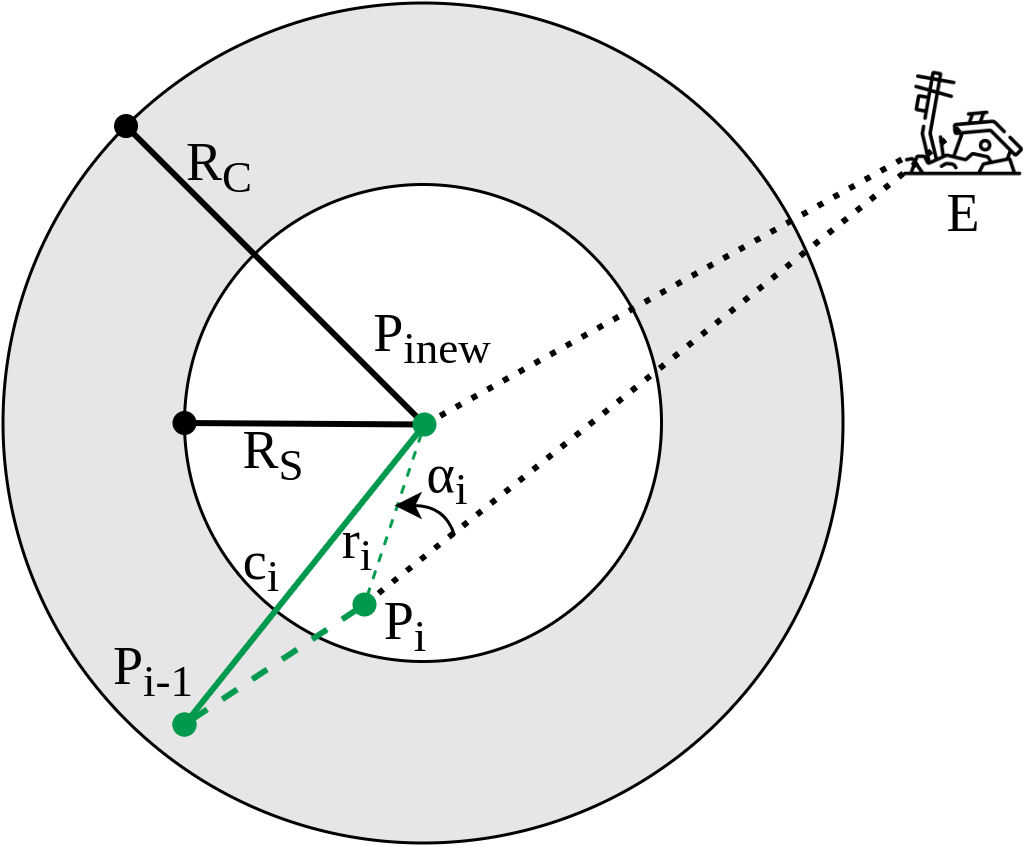}
    \caption{Illustration of the target angle $\alpha$}
    \label{fig:Rel2Point}
\end{figure}

Assume the disaster area is known and its center $E$ is the target location for UAVs' deployment. The target angle $\alpha_i$ is then defined as the angle between the vector $\overrightarrow{P_{i}P_{inew}}$ from the current position $P_i$ to the next search position $P_{inew}$ and the vector $\overrightarrow{P_{i}E}$ towards the target as illustrated in Fig. \ref{fig:Rel2Point}. The next position to deploy the UAV is expected to have the smallest target angle, i.e., closest to the line of sight (LOS) between the base station and the target. Thus, the objective function to minimize the target angle is defined as follows:
\begin{equation}
    F_2=\alpha_{i}=\text{atan2}\left({\left\Vert \overrightarrow{P_{i}P_{inew}}\times\overrightarrow{P_{i}E}\right\Vert ,\overrightarrow{P_{i}P_{inew}}\cdot\overrightarrow{P_{i}E}}\right)
\end{equation}
where operators `$\times$' and `$\cdot$' are the cross and dot products, respectively.

\subsubsection{Optimizing communication links}
During the deployment process, the communication link between two adjacent UAVs needs to be optimized to ensure communication quality and minimize the number of UAVs used. For that, the location of the next UAV must be within the communication range of its adjacent ones.

Let $P_i$, $P_{inew}$, and $P_{i-1}$ be respectively the positions of the current UAV, the optimal point, and the previous adjacent node. Denote $c_i$ and $r_i$ as the distances between $P_{i-1}$ and $P_{i}$, and $P_i$ and $P_{inew}$, respectively. The objective functions to optimize those distances are defined as follows:

\begin{equation}
    F_{3}=\left\{ \begin{array}{cc}
    \left\vert R_S-r_{i}\right\vert &\text{if } r_{i}\leq R_{S}\\
    \infty &\text{otherwise}
    \end{array}\right.
\end{equation}
\begin{equation}
    F_{4}=\left\{ \begin{array}{cc}
    \left\vert R_C-c_{i}\right\vert &\text{if } c_{i}\leq R_{C}\\
    \infty &\text{otherwise}
    \end{array}\right.
\end{equation}



\subsection{Optimal location identification using PSO}
The PSO algorithm is used to minimize $F$ to find the best positions for UAVs due to its fast convergence and robustness in solving complex optimization problems \cite{poli2007particle,PHUNG2021107376}. It works based on swarm intelligence in which a set of particles is first initiated with random positions. Their positions are then adjusted based on two learning parameters, $gbest$, the best position of the swarm, and $pbest_i$, the best position of particle $i$. Let $v_i(t)$ be the velocity of particle $i$ at time $t$. Its position $x_i(t)$ is then updated according to the following equations:
\begin{equation}
    \begin{aligned}
    v_{i}(t+1)&=\text{\ensuremath{\omega}}v_{i}(t)+c_{1}r_{1}(pbest_{i}(t)-x_{i}(t))\\
    &+c_{2}r_{2}(gbest(t)-x_{i}(t)) \\
    x_{i}(t+1) &= x_{i}(t)+v_{i}(t+1)
    \end{aligned}
    \label{eqn:pso}
\end{equation}
where $\omega$ is the inertia coefficient; $c_1$, $c_2$ are respectively the individual and global acceleration coefficients; and $r_1$, $r_2$ are two random values in the range of $(0,1)$. 

\begin{algorithm}
\KwResult{Optimal position}
Initialize: PSO population, $gbest$, and $IterMax$\\
Observe obstacles in environment\\
\For{$k\leftarrow 1$ to $IterMax$}
{
    \For {Particle $i$ in PSO Population}
    {
        Calculate cost value $F$ of Particle $i$ by Eq. \eqref{eqn:cost}\\
        Evaluate and Update $pbest_i$\\ 
        Evaluate and Update $gbest$\\
        Update velocities $v_i$ and positions $x_i$ by Eq. \eqref{eqn:pso}\\
    }
}
\caption{PSO-based optimal location identification}
\label{alg:pso}
\end{algorithm}

The implementation of PSO to find UAVs' optimal position is described in Algorithm \ref{alg:pso}. Initially, a set of particles is generated with random positions, each represents a possible position of the UAV. Obstacles in the searching area are also identified by the UAV's sensor. The searching process then starts in which each particle $i$ calculates its fitness value using Eq. \eqref{eqn:cost}. The best cost values of particle $i$, $pbest_i$, and of the swarm, $gbest$, are then updated. Finally, the velocity $v_i$ and position $x_i$ of particle $i$ are updated by Eq. \eqref{eqn:pso}. The process is repeated until the maximum number of iterations is reached. The position associated with the final $gbest$ value is chosen as the optimal position for the UAV. 

\section{Behavior-based Target Tracking}
\label{sec:tra}
\begin{figure*}[!t]
    \centering
    \begin{subfigure}[b]{0.22\textwidth}
    \centering
    \includegraphics[width=\textwidth]{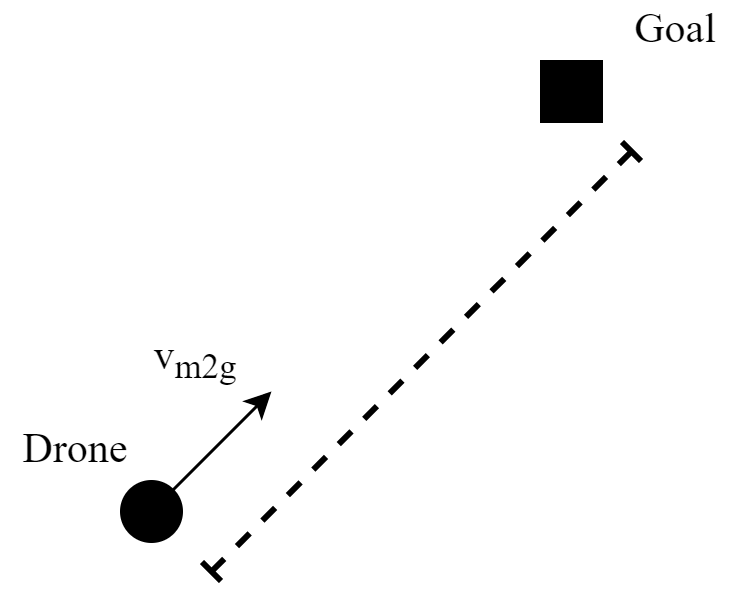}
    \caption{Move to the target}
    \label{fig:m2g}
    \end{subfigure}
    \hspace{1.5cm}
    \begin{subfigure}[b]{0.2\textwidth}
    \centering
    \includegraphics[width=\textwidth]{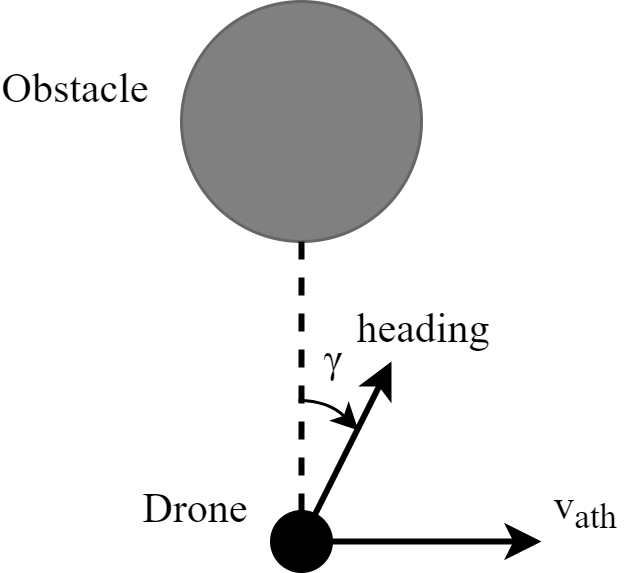}
    \caption{Avoid obstacles}
    \label{fig:ath}
    \end{subfigure}
    \hspace{1.5cm}
    \begin{subfigure}[b]{0.2\textwidth}
    \centering
    \includegraphics[width=\textwidth]{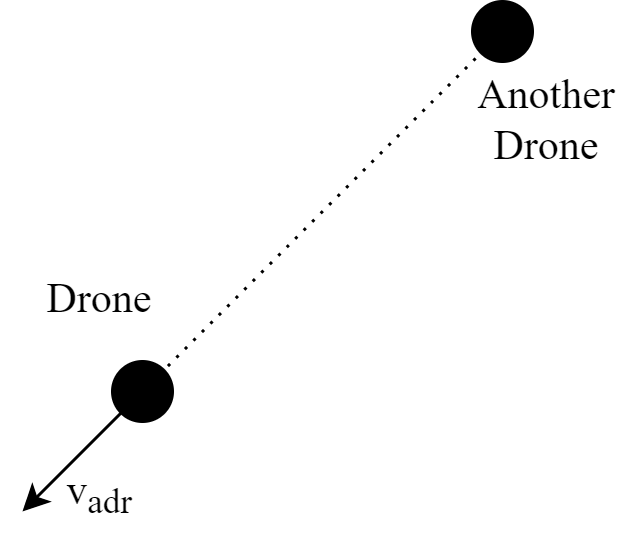}
    \caption{Avoid other UAVs}
    \label{fig:adr}
    \end{subfigure}
    \caption{Sub-behaviors of UAVs}
    \vspace{-0.5cm}
    \label{fig:sub}
\end{figure*}

After finding the optimal position $P_i$ for UAV $i$, the deployment of the  UAV continues by first navigating it to $P_i$ and then repeating the search process to find the optimal position $P_{i+1}$. To navigate the UAV to $P_i$, we design a behavior-based target tracking controller that combines several sub-behavior of the form:

\begin{equation}
    V = f_{m2g}V_{m2g} + f_{ath}V_{ath} + f_{adr}V_{adr},
    \label{eqn:v}
\end{equation}
where $V_{m2g}$, $V_{ath}$, $V_{adr}$ are respectively the sub-behaviors for moving to the target, avoiding obstacles, and avoiding other UAVs, and $f_{m2g}$, $f_{ath}$, $f_{adr}$ are control parameters. The definition of sub-behavior is described as follows.

\subsection{Move-to-target behavior}
The aim of the move-to-target behavior is to drive the UAV toward its target position as illustrated in Fig. \ref{fig:m2g}. Denote the target position as $P_d=\left[x_d, y_d, z_d\right]^T$. The move-to-target behavior is defined as:
\begin{equation}
    V_{m2g}=\dfrac{1}{d_{m2g}}\left[\begin{array}{c}
    x_{d}-x\\
    y_{d}-y\\
    z_{d}-z
    \end{array}\right]
\end{equation}
where $d_{m2g}$ is the distance from the UAV to its target $P_d$. The control parameter for this behavior is defined as follows:
\begin{equation}
    f_{m2g}\left(d_{m2g}\right)=\left\{ \begin{array}{cc}
    a_{m2g} & \text{if } d_{m2g}\geq b_{m2g}\\
    a_{m2g}\dfrac{d_{m2g}}{b_{m2g}} & \text{if } d_{m2g}<b_{m2g}
    \end{array}\right.
\end{equation}
where $a_{m2g}$ and $b_{m2g}$ are pre-defined constants.

\subsection{Obstacle avoidance behavior}
When an obstacle is detected, the UAV activates its avoidance behavior. With this behavior, the direction of the UAV's movement is changed by $90$ degrees relative to the direction towards the obstacle and the rotation angle $\gamma$ is changed depending on the direction of the target as illustrated in Fig. \ref{fig:ath}. Assume that at height $z$, the obstacle is modeled as a circle with center $\left(x_{th},y_{th}\right)$ and radius $r_{th}$. The obstacle avoidance behavior is then expressed as:
\begin{equation}
    V_{ath}=\dfrac{1}{d_{ath}}\text{Rot}_z\left(-\dfrac{\pi}{2}\text{sgn}\left(\rho\right)\right)\left[\begin{array}{c}
    x_{th}-x\\
    y_{th}-y\\
    z
    \end{array}\right]
\end{equation}
where $d_{ath}$ is the distance from the UAV to the obstacle, Rot$_z$ represents the rotation about the $z$-axis, and sgn(.) denotes the sign function. The rotation angle $\gamma$ is determined as follows:
\begin{equation}
    \rho=\sin\gamma=\dfrac{x_{th}-x}{d_{ath}}\sin\psi-\dfrac{y_{th}-y}{d_{ath}}\cos\psi
\end{equation}
The control parameter for the obstacle avoidance behavior is given as follows:
\begin{equation}
    f_{ath}\left(d_{ath}\right)=\left\{ \begin{array}{cc}
    0 & \text{if } d_{ath}\geq b_{ath}\\
    a_{ath}\left(1-\dfrac{d_{ath}}{b_{ath}}\right) & \text{if } d_{ath}<b_{ath}
    \end{array}\right.
\end{equation}
where $a_{ath}$ and $b_{ath}$ are the adjustable parameters.

\subsection{Avoiding collision with other UAVs}
The behavior to avoid collision with other UAVs is activated when one UAV is too close to another. As illustrated in Fig. \ref{fig:adr}, the avoidance is carried out based on control vector $V_{adr}$ that drives  the UAV to move away from the other. Let $d_{adr}$ be the distance between two UAVs, and $P_{dr}=\left[x_{dr},y_{dr}, z_{dr}\right]^T$ be the position of the UAV to be avoided. The collision avoidance behavior is formulated as follows:
\begin{equation}
    V_{adr}=\dfrac{1}{d_{adr}}\left[\begin{array}{c}
    x_{dr}-x\\
    y_{dr}-y\\
    z_{dr}-z
    \end{array}\right]
\end{equation}
The control parameter of this behavior is given as:
\begin{equation}
    f_{adr}\left(d_{adr}\right)=\left\{ \begin{array}{cc}
    0 & \text{if } d_{adr}\geq b_{adr}\\
    a_{adr}\left(1-\dfrac{d_{adr}}{b_{adr}}\right) & \text{if } d_{adr}<b_{adr}
    \end{array}\right.
\end{equation}
where $a_{adr}$ and $b_{adr}$ are pre-defined constants.

\section{Multihop Ad-hoc Route Deployment}
\label{sec:sys}

\begin{algorithm}  
\Switch{state}
{
    \Case{Unassigned}{
        Be in idle\\
        \If {be its turn and get target}
        {
            state $\leftarrow$ Assigned\\
        }
    }
    
    \Case{Assigned}{
        Get $current\_node$, $target\_node$\\
        \If {($current\_node$ != $target\_node$)}
        {
            Move to $current\_node$ by Behavior-based controller\\
            $current\_node$ = $next\_node$
        }
        \Else {state $\leftarrow$ Explore}
    }
    \Case{Explore}{
        
        \If {Not existing Target}
        {
            Target searching using PSO (Algorithm \ref{alg:pso})\\
            \If{ Reach limited range or destination} 
            {
                state $\leftarrow$ Occupied
            }
        }
        \Else
        {
            Move to $target$ by Behavior-based controller\\
            \If {Reach target}
            {
                Hover\\
            }
        }
    }
    \Case{Occupied}{
        \eIf {Reach destination}
        {
            Multihop ad-hoc route is done
        }
        {
            Broadcast current location until assigned\\
        }
    }
}
\caption{Multihop Ad-hoc Communication Route Deployment Strategy}
\label{alg:strategy}
\end{algorithm}

In this section, we present the strategy to deploy UAVs to form a multihop ad-hoc communication network from the base station to the disaster area using the PSO and behavior-based controller\footnote{Source code of the strategy can be found at {\fontfamily{qcr}\selectfont \url{https://github.com/duynamrcv/uav_multihop_adhoc}}}. As shown in Algorithm \ref{alg:strategy}, the strategy uses four states named unassigned, assigned, explore, and occupied as in \cite{Hung7576911} to deploy each UAV as follows.

\textbf{``Unassigned''} (lines 2-7): the UAV is idle at the base station and waits to be deployed. If it receives the position and in-turn messages from the current deploying UAV, it changes its state to ``Assigned''.

\textbf{``Assigned''} (lines 8-17): the UAV is assigned a target which is the optimal position found by the last UAV. It then flies to the target using the behavior-based controller. During the flight, the UAV needs to go through the established nodes. When reaching the target, its state is changed to ``Explore''.

\textbf{``Explore''} (lines 18-31): the UAV explores the next optimal intermediate position within its sensing range using the PSO algorithm. It then repeats the exploration process until reaching its communication range. The last optimal intermediate position is set as the final position of the UAV. The UAV hovers at that position and changes its state to ``Occupied''.

\textbf{``Occupied''} (lines 32-38): the UAV is hovering at its optimal position. If that position is not the destination, the UAV will broadcast its position to the unassigned UAVs until a UAV is dispatched. Otherwise, the establishment of the multihop ad-hoc route is completed.

\begin{table*}
\centering
\caption{Results of the UAV deployment compared to the ideal reference}
\label{tbl:res}
\begin{tabular}{|c|c|c|c|c|c|c|c|c|c|} 
\hline
\multicolumn{1}{|c|}{\multirow{3}{*}{Scen.}} & \multicolumn{1}{c|}{\multirow{3}{*}{UAV}} & \multicolumn{1}{c|}{\multirow{3}{*}{Ideal position}} & \multicolumn{1}{c|}{\multirow{3}{*}{Actual position}} & \multicolumn{2}{c|}{Target Angle (rad)}                                     & \multicolumn{2}{c|}{Time consuming (s)}                                           & \multicolumn{1}{c|}{\multirow{3}{*}{\begin{tabular}[c]{@{}c@{}}Link \\ length \\ (m)\end{tabular}}} & \multicolumn{1}{c|}{\multirow{3}{*}{\begin{tabular}[c]{@{}c@{}} Temporary \\ goal \\ number\end{tabular}}} \\ \cline{5-8}
\multicolumn{1}{|c|}{}                       & \multicolumn{1}{c|}{}                     & \multicolumn{1}{c|}{}                                & \multicolumn{1}{c|}{}                                 & \multicolumn{1}{l|}{\multirow{2}{*}{Ideal}} & \multirow{2}{*}{Actual} & \multicolumn{1}{l|}{\multirow{2}{*}{Searching}} & \multirow{2}{*}{Deployment} & \multicolumn{1}{c|}{}                                                                               & \multicolumn{1}{c|}{}                                                                                   \\
\multicolumn{1}{|c|}{}                       & \multicolumn{1}{c|}{}                     & \multicolumn{1}{c|}{}                                & \multicolumn{1}{c|}{}                                 & \multicolumn{1}{l|}{}                       &                         & \multicolumn{1}{l|}{}                           &                             & \multicolumn{1}{c|}{}                                                                               & \multicolumn{1}{c|}{}                                                                                   \\ \hline
\multirow{4}{*}{1}     & 1                    & $\left[405.92,318.04,323.47\right]^T$ & $\left[430.38,280.78,328.26\right]^T$ & \multirow{4}{*}{0.8140} & 0.6653  & 12.05   & 21.05                       & 293.06                           & 7                               \\ 
\cline{2-4}\cline{6-10}
                       & 2                    & $\left[611.85,536.07,330.98\right]^T$ & $\left[669.47,452.65,367.33\right]^T$ &                         & 0.6233  & 15.21   & 46.19                       & 297.04                           & 9                               \\ 
\cline{2-4}\cline{6-10}
                       & 3                    & $\left[817.77,754.11,338.49\right]^T$ & $\left[818.47,706.88,351.12\right]^T$ &                         & 1.0407  & 11.76  & 60.45                       & 295.12                           & 7                               \\ 
\cline{2-4}\cline{6-10}
                       & 4                    & $\left[880,820,340.70\right]^T$ & $\left[878.70,817.88,344.90\right]^T$ &                         & 1.0737  & 5.47       & 69.76                       & 126.44                           & 4                              \\ 
\hline
\multirow{4}{*}{2}     & 1                    & $\left[387.38,334.23,332.65\right]^T$ & $\left[357.74,349.55,328.01\right]^T$ & \multirow{4}{*}{0.8961} & 1.0071  & 13.17   & 20.91                       & 295.44                           & 7                               \\ 
\cline{2-4}\cline{6-10}
                       & 2                    & $\left[574.77,568.46,329.35\right]^T$ & $\left[584.39,539.50,322.09\right]^T$ &                         & 0.6975  & 15.41   & 42.94                       & 295.78                           & 8                               \\ 
\cline{2-4}\cline{6-10}
                       & 3                    & $\left[762.15,802.69,336.04\right]^T$ & $\left[798.24,741.52,350.71\right]^T$ &                         & 0.7570  & 14.89  & 63.21                       & 295.57                           & 8                               \\ 
\cline{2-4}\cline{6-10}
                       & 4                    & $\left[800,850,337.39\right]^T$ & $\left[799.29,845.82,339.34\right]^T$ &                         & 1.5607  & 4.031       & 68.00                       & 104.92                           & 3                              \\
\hline
\end{tabular}
\end{table*}

\section{Results}



\begin{figure*}
\centering
    \begin{subfigure}[b]{0.44\textwidth}
    \centering
    \includegraphics[width=\textwidth]{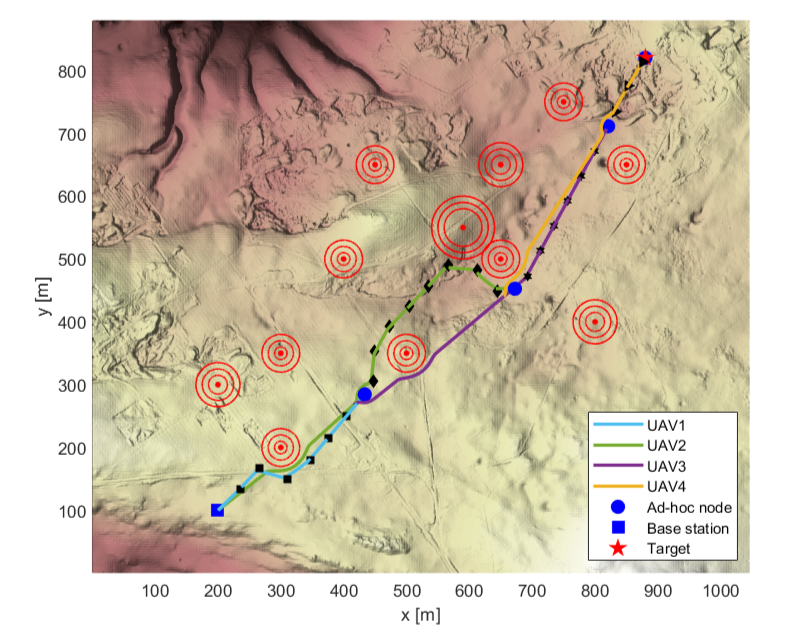}
    \caption{Scenario 1}
    \label{fig:top1}
    \end{subfigure}
    \begin{subfigure}[b]{0.44\textwidth}
    \centering
    \includegraphics[width=\textwidth]{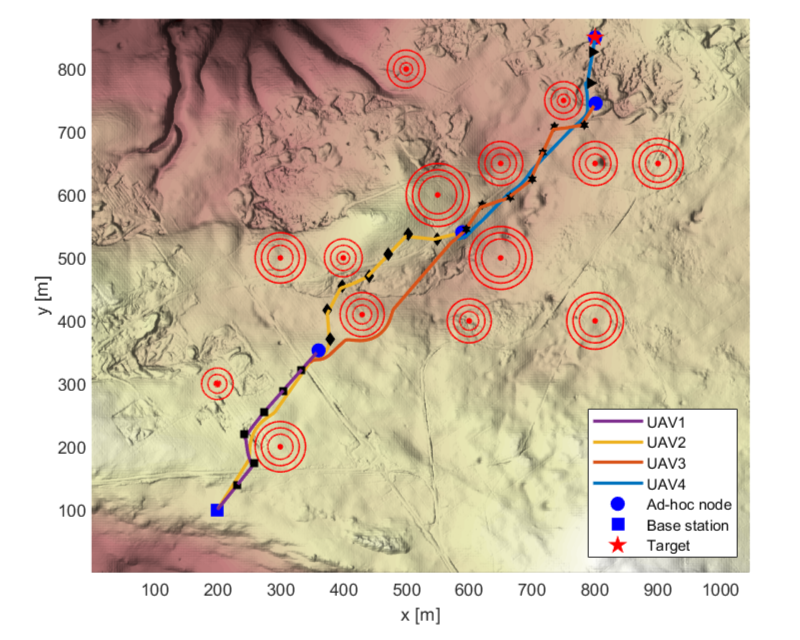}
    \caption{Scenario 2}
    \label{fig:top2}
    \end{subfigure}
    \caption{Results of the UAV deployment for multihop ad-hoc communication routes}
    \label{fig:top}
\end{figure*}

To evaluate the performance of the proposed approach, we have carried out a number of simulation experiments. Two scenarios are generated by using the digital elevation model (DEM) map \cite{https://doi.org/10.26186/89644} augmented with additional obstacles as shown in Fig. \ref{fig:top}. The distance between the base station and the destination is chosen as 1000 m in both scenarios. The UAV has $D=1$ m, $S=10D$, $R_C=300$ m, and $R_S=50$ m. The PSO is set with the population size of 100, $w=1$ with a damping ratio of $0.98$, $c_1=c_2=1.5$, and $IterMax=100$. The behavior-based target tracking control is set with $a_{m2g}=30.0$, $b_{m2g}=20.0$, $a_{ath} = 30.0$, $b_{ath} =15.0$, $a_{adr}=30.0$ and $b_{adr}=20.0$. With this setup, the system requires 4 UAVs to form a multihop ad-hoc route. 

Figure \ref{fig:top} shows the deployment results\footnote{Video showing the proposed strategy deployment process: {\fontfamily{qcr}\selectfont \url{https://youtu.be/5w8bz2UjB94}}}. In both scenarios, $4$ UAVs have been successfully deployed as ad-hoc nodes (blue circles) located evenly along the route with an average link length of 295 m (excluding the link to the destination), which is close to the communication range of 300 m.  The trajectory of each UAV consists of two segments corresponding to its operation in the ``Assigned'' and ``Explore'' states. Both segments have safe distances from obstacles. The first segment is the result of the behavior-based controller when the UAV flies along the route to reach the last node. The second segment is obtained by the PSO when it guides the UAV through intermediate points, marked by black dots, until reaching the communication boundary. The number of intermediate points depend on the ratio between the sensing and the communication ranges. Noting that the paths generated by the behavior-based controller need to go through the nodes occupied by the UAVs (blue circles), but not the intermediate points (black dots) since those points are found during the exploration process and maybe not globally optimal.

\begin{figure*}
\centering
    \begin{subfigure}[b]{0.44\textwidth}
    \centering
    \includegraphics[width=\textwidth]{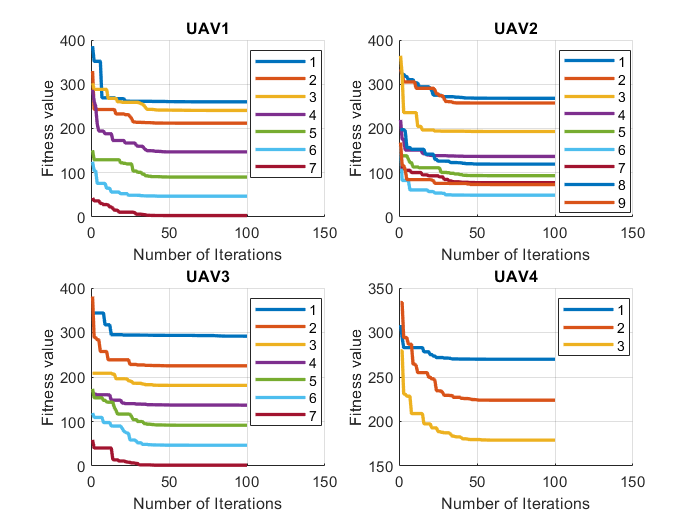}
    \caption{Scenario 1}
    \end{subfigure}
    \begin{subfigure}[b]{0.44\textwidth}
    \centering
    \includegraphics[width=\textwidth]{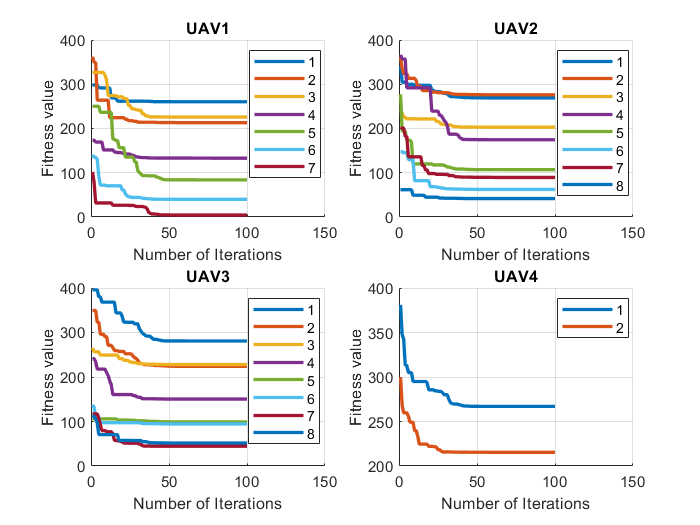}
    \caption{Scenario 2}
    \end{subfigure}
    \caption{Convergence rate of the PSO-based optimal position searching algorithm}
    \label{fig:iter}
\end{figure*}

Fig. \ref{fig:iter} shows the fitness values over iterations when the PSO looks for optimal positions. It can be seen that the algorithm quickly converges after 30 to 50 iterations. In addition, the fitness values decrease when the UAV approaches its final optimal position. Table \ref{tbl:res} shows the data obtained from two simulation experiments in which the ideal position and target angle are determined along the line of sight between the base station and destination. The data shows that the actual positions are close to their ideal references with an average deviation of 70 m. The actual target angles are also close to their ideal values with an average deviation of 0.2 rad. The deviations are mainly caused by obstacles appeared in the environment. 

\label{sec:res}

In terms of time, the system requires an average time of 5 minutes to complete the deployment at the UAV's velocity of 15 m/s. The searching for an optimal point using PSO takes about 1 s, which is sufficient since the UAV will take at least 3.3 s to fly from one point to another due to its 50 m sensing range. The behavior-based controller only requires milliseconds to compute the control command due to its simplicity in computation. The proposed algorithms therefore can be used for real-time deployment of UAVs. They are also scalable for larger distances since the computation does not depend on the sensing or communication range.
\section{Conclusion}
\label{sec:con}
In this work, we have presented a new approach to form a multihop ad-hoc route to provide communication for disaster response scenarios using a distributed UAV system. The proposed approach is a combination of a searching method based on the PSO algorithm to find optimal positions to place network nodes and a distributed behavior-based controller to coordinate the UAVs to move along the established route to reach their desired position while avoiding collisions with obstacles and other UAVs. The evaluation results show that the proposed approach can establish ad-hoc networks in complex environments with many obstacles. The deployed positions of the UAVs and the flight paths are close to their ideal values. The computation time is also sufficient for real-time operation. Those results confirm the efficiency and applicability of our method for real-world scenarios.

\bibliographystyle{IEEEtran}  
\bibliography{ref}  

\begin{thebibliography}{10}
\providecommand{\url}[1]{#1}
\csname url@samestyle\endcsname
\providecommand{\newblock}{\relax}
\providecommand{\bibinfo}[2]{#2}
\providecommand{\BIBentrySTDinterwordspacing}{\spaceskip=0pt\relax}
\providecommand{\BIBentryALTinterwordstretchfactor}{4}
\providecommand{\BIBentryALTinterwordspacing}{\spaceskip=\fontdimen2\font plus
\BIBentryALTinterwordstretchfactor\fontdimen3\font minus
  \fontdimen4\font\relax}
\providecommand{\BIBforeignlanguage}[2]{{%
\expandafter\ifx\csname l@#1\endcsname\relax
\typeout{** WARNING: IEEEtran.bst: No hyphenation pattern has been}%
\typeout{** loaded for the language `#1'. Using the pattern for}%
\typeout{** the default language instead.}%
\else
\language=\csname l@#1\endcsname
\fi
#2}}
\providecommand{\BIBdecl}{\relax}
\BIBdecl

\bibitem{doi:10.1155/2015/647037}
D.~G. Reina, M.~Askalani, S.~L. Toral, F.~Barrero, E.~Asimakopoulou, and
  N.~Bessis, ``A survey on multihop ad hoc networks for disaster response
  scenarios,'' \emph{International Journal of Distributed Sensor Networks},
  vol.~11, no.~10, p. 647037, 2015.

\bibitem{6710069}
M.~Conti and S.~Giordano, ``Mobile ad hoc networking: milestones, challenges,
  and new research directions,'' \emph{IEEE Communications Magazine}, vol.~52,
  no.~1, pp. 85--96, 2014.

\bibitem{anjum2017review}
S.~S. Anjum, R.~M. Noor, and M.~H. Anisi, ``Review on manet based communication
  for search and rescue operations,'' \emph{Wireless personal communications},
  vol.~94, no.~1, pp. 31--52, 2017.

\bibitem{5730373}
T.~D. Ta, T.~D. Tran, D.~D. Do, H.~V. Nguyen, Y.~V. Vu, and N.~X. Tran,
  ``Gps-based wireless ad hoc network for marine monitoring, search and rescue
  (msnr),'' in \emph{2011 Second International Conference on Intelligent
  Systems, Modelling and Simulation}, 2011, pp. 350--354.

\bibitem{8167124}
V.~Tundjungsari and A.~Sabiq, ``Android-based application using mobile adhoc
  network for search and rescue operation during disaster,'' in \emph{2017
  International Conference on Electrical Engineering and Computer Science
  (ICECOS)}, 2017, pp. 16--21.

\bibitem{8756125}
V.~T. Hoang, M.~D. Phung, T.~H. Dinh, and Q.~P. Ha, ``System architecture for
  real-time surface inspection using multiple {UAVs},'' \emph{IEEE Systems
  Journal}, vol.~14, no.~2, pp. 2925--2936, 2020.

\bibitem{9214446}
Y.~Zhou, B.~Rao, and W.~Wang, ``Uav swarm intelligence: Recent advances and
  future trends,'' \emph{IEEE Access}, vol.~8, 2020.

\bibitem{PHUNG2020106705}
M.~D. Phung and Q.~P. Ha, ``Motion-encoded particle swarm optimization for
  moving target search using {UAVs},'' \emph{Applied Soft Computing}, vol.~97,
  p. 106705, 2020.

\bibitem{maxa2015secure}
J.-A. Maxa, M.~S.~B. Mahmoud, and N.~Larrieu, ``Secure routing protocol design
  for uav ad hoc networks,'' in \emph{2015 IEEE/AIAA 34th Digital Avionics
  Systems Conference (DASC)}.\hskip 1em plus 0.5em minus 0.4em\relax IEEE,
  2015, pp. 4A5--1.

\bibitem{sun2020data}
J.~Sun, W.~Wang, L.~Kou, Y.~Lin, L.~Zhang, Q.~Da, and L.~Chen, ``A data
  authentication scheme for uav ad hoc network communication,'' \emph{The
  Journal of Supercomputing}, vol.~76, no.~6, pp. 4041--4056, 2020.

\bibitem{zhao2012topology}
Z.~Zhao and T.~Braun, ``Topology control and mobility strategy for uav ad-hoc
  networks: A survey,'' in \emph{Joint ERCIM eMobility and MobiSense
  workshop}.\hskip 1em plus 0.5em minus 0.4em\relax Citeseer, 2012, pp. 27--32.

\bibitem{chua2012medium}
M.~Y.-K. Chua, F.~R. Yu, J.~Li, Y.~Zhou, and L.~Lamont, ``Medium access control
  for unmanned aerial vehicle (uav) ad-hoc networks with full-duplex radios and
  multipacket reception capability,'' \emph{IEEE Transactions on Vehicular
  Technology}, vol.~62, no.~1, pp. 390--394, 2012.

\bibitem{poli2007particle}
R.~Poli, J.~Kennedy, and T.~Blackwell, ``Particle swarm optimization,''
  \emph{Swarm intelligence}, vol.~1, no.~1, pp. 33--57, 2007.

\bibitem{PHUNG2021107376}
M.~D. Phung and Q.~P. Ha, ``Safety-enhanced {UAV} path planning with spherical
  vector-based particle swarm optimization,'' \emph{Applied Soft Computing},
  vol. 107, p. 107376, 2021.

\bibitem{Hung7576911}
P.~D. Hung, T.~Q. Vinh, and T.~D. Ngo, ``Distributed coverage control for
  networked multi-robot systems in any environments,'' in \emph{2016 IEEE
  International Conference on Advanced Intelligent Mechatronics (AIM)}, 2016,
  pp. 1067--1072.

\bibitem{https://doi.org/10.26186/89644}
{Geoscience Australia}, ``Digital elevation model (dem) of australia derived
  from lidar 5 metre grid,'' 2015.

\end{thebibliography}
\balance

\end{document}